\pdfoutput=1

\documentclass[11pt]{article}

\usepackage[final]{acl}

\usepackage{times}
\usepackage{latexsym}
\usepackage{makecell}
\usepackage{float}

\usepackage[T1]{fontenc}

\usepackage[utf8]{inputenc}

\usepackage{microtype}
\usepackage{amsmath}

\usepackage{inconsolata}

\usepackage{graphicx}

\usepackage{booktabs}
\usepackage{multirow}
\usepackage[capitalise]{cleveref}
\crefformat{section}{\S#2#1#3} 
\crefformat{subsection}{\S#2#1#3}
\crefformat{subsubsection}{\S#2#1#3}
\usepackage{todonotes}
\usepackage{subcaption}

\title{Plane Geometry Problem Solving with Multi-modal Reasoning: A Survey}

\author{
 \textbf{Seunghyuk Cho\textsuperscript{1}},
 \textbf{Zhenyue Qin\textsuperscript{3}},
 \textbf{Yang Liu\textsuperscript{3}},
 \textbf{Youngbin Choi\textsuperscript{1}},
\\
 \textbf{Seungbeom Lee\textsuperscript{1}},
 \textbf{Dongwoo Kim\textsuperscript{1,2}}
\\
 \textsuperscript{1}Graduate School of Artificial Intelligence, POSTECH,
\\
 \textsuperscript{2}Department of Computer Science and Engineering, POSTECH, 
\\
 \textsuperscript{3}Australian National University
\\
 \small{
   \textbf{Correspondence to:} Dongwoo Kim \href{mailto:dongwoo.kim@postech.ac.kr}{<dongwoo.kim@postech.ac.kr>}
 }
}

\begin{document}
\maketitle
\begin{abstract}
Plane geometry problem solving (PGPS) has recently gained significant attention as a benchmark to assess the multi-modal reasoning capabilities of large vision-language models. 
Despite the growing interest in PGPS, the research community still lacks a comprehensive overview that systematically synthesizes recent work in PGPS. 
To fill this gap, we present a survey of existing PGPS studies.
We first categorize PGPS methods into an encoder-decoder framework and summarize the corresponding output formats used by their encoders and decoders. 
Subsequently, we classify and analyze these encoders and decoders according to their architectural designs.
Finally, we outline major challenges and promising directions for future research. 
In particular, we discuss the hallucination issues arising during the encoding phase within encoder-decoder architectures, as well as the problem of data leakage in current PGPS benchmarks.
\end{abstract}

\section{Introduction}

Automated plane geometry problem solving (PGPS) has emerged as an important benchmark in artificial intelligence research due to its unique requirement for multi-modal reasoning with mathematical rigor~\citep{geos, geoqa}. Typically, geometry problems combine textual descriptions with visual diagrams, each providing essential complementary information. The inherent necessity to integrate linguistic and visual modalities makes plane geometry a compelling testbed for advancing the multi-modal understanding capabilities of AI systems. Furthermore, practical motivations such as developing intelligent tutoring systems~\citep{tutor1, tutor2, tutor3} and standardized benchmarks for evaluating AI reasoning~\citep{geoqa, geoqa+} highlight the importance of continued research in this area.

Nevertheless, substantial challenges persist in achieving full automation. Foremost among these is the complexity arising from the multi-modal nature of geometry problems, requiring precise alignment between textual statements and corresponding diagram elements~\citep{G-Aligner}. 
Resolving ambiguities in textual descriptions through visual references and accurately mapping entities between text and diagrams pose significant hurdles~\citep{geos++,pgdp5k}. 
Geometric diagrams also introduce unique challenges absent in natural images and other types of diagrams, including precise recognition of abstract symbols, e.g., angle markers and length indicators, accurate detection of geometric primitives, e.g., points, lines, and circles, and interpretation of implicit spatial relationships governed by geometric constraints. Additionally, effective PGPS demands embedding deep geometric domain knowledge, applying geometric axioms and theorems during the reasoning that are often implicitly assumed~\citep{geos++,GEOS-OS,intergps}. Thus, integrating linguistic comprehension, visual analysis, and geometric reasoning continues to drive the complexity and significance of research in automated PGPS.

Recently, numerous new benchmarks, large-scale datasets, and model architectures have been proposed to tackle the challenges of PGPS.
However, despite this rapid progress, most existing surveys on mathematical or multi-modal reasoning address geometry problems only as part of broader domains~\citep{survey1,survey2,survey3} and thus fail to examine the unique challenges of PGPS in depth.
Consequently, the literature still lacks a dedicated, up-to-date survey centered on PGPS. 
The goal of this paper is to fill the gap by providing the PGPS research community with a structured overview of the latest benchmarks, datasets, and multi-modal reasoning approaches tailored specifically to PGPS.

The structure of this paper is summarized as follows: We first describe the definition of PGPS and relevant tasks~(\cref{sec:tasks}). We then introduce an overall framework for solving PGPS problems as an encoder-decoder architecture with intermediate representations~(\cref{sec:overall}). Next, we review the details of encoder~(\cref{sec:encoder}) and decoder~(\cref{sec:decoder}) structures. Some additional thoughts are provided from the data collection perspective~(\cref{sec:misc}). Finally, we address the remaining challenges and promising future directions in automated PGPS~(\cref{sec:challenges}).

\begin{figure*}[t!]
    \centering
    \includegraphics[width=.8\linewidth]{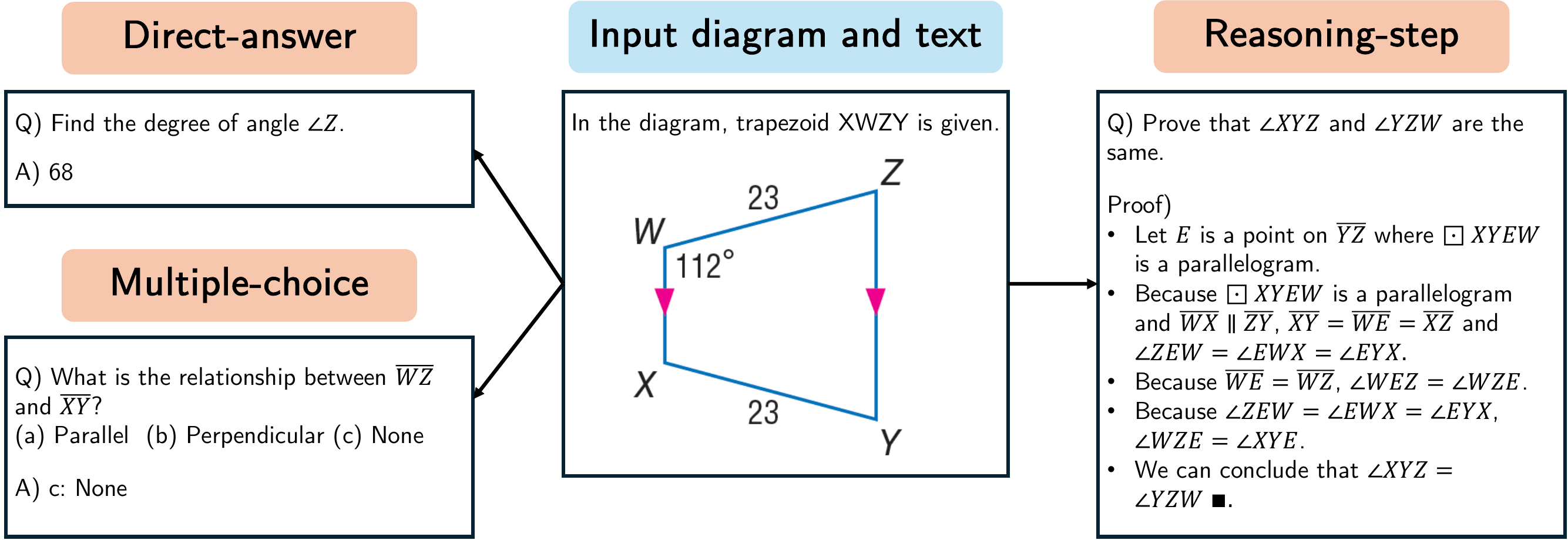}
    \caption{Illustration of three PGPS tasks. The three tasks are commonly used to evaluate PGPS methods in existing benchmarks: i) direct-answer, ii) multiple-choice, and iii) reasoning-step construction. In the direct-answer task, the model must predict a single numerical value as the answer to the problem. In the multiple-choice task, the model must select the correct label corresponding to the ground-truth option. In the reasoning-step construction task, the model is asked to generate the complete sequence of reasoning steps that lead to the correct final answer.}
    \label{fig:pgps_tasks}
\end{figure*}

\section{Tasks and benchmarks}\label{sec:tasks}
In this section, we first define the PGPS and then introduce three tasks that are commonly tackled in the PGPS community, along with the benchmarks for each task.  

\subsection{Definition of PGPS}

Euclidean plane geometry studies the properties and relationships among geometric primitives, e.g., points, lines, and circles, in a flat, two-dimensional space~\citep{fitzpatrick2007euclid}. 
PGPS involves inferring unknown geometric properties or relationships from a given set of primitives and their known relations, such as determining the length of an unknown side in a triangle given the lengths of two sides and the measure of the included angle.

In real-world scenarios, plane geometry problems usually present as diagram and textual description pairs, as demonstrated in \cref{fig:pgps_tasks}.
The diagrams and accompanying textual descriptions typically complement each other in representing geometric primitives and relations. Diagrams usually provide visual information about spatial relationships, whereas textual descriptions explicitly mention properties or relational details. Due to this complementary nature, PGPS methods in real-world applications must not only infer unknown geometric facts but also accurately parse geometric information from these diagrams and text pairs.

\subsection{PGPS tasks}

We describe the three main tasks, along with the corresponding benchmarks, that are mainly tackled via PGPS research. \cref{fig:pgps_tasks} illustrates three examples for each task.
For further details on the benchmarks from various perspectives, such as reasoning complexity, diagram-text interdependency, and data collection methods, refer to \cref{sec:misc}.

\subsubsection{Direct-answer and multiple-choice tasks}

\paragraph{Task description}
Most PGPS works quantify the capacity of a PGPS method to infer a single, well-defined property of a geometric entity from a unified diagrammatic–textual problem statement. The requested properties fall into two categories: i) numerical targets, e.g., angle magnitude, segment length, or area~\citep{geos,intergps,geoqa}, and ii) categorical targets, e.g., the perpendicularity or parallelism of two lines~\citep{geosense}.

PGPS methods are also evaluated through multiple-choice tasks~\citep{mathvista,mathverse}. While these tasks use the same problems as direct-answer tasks, each multiple-choice problem provides a fixed set of candidate responses. A PGPS method must select the option that correctly identifies the target property, or equivalently, predict a value matching one of the provided choices. For example, in the scenario depicted in \cref{fig:pgps_tasks}, the correct response is the label "c" or its corresponding value, "None."

\paragraph{Evaluation metrics}
In direct-answer tasks, performance is reported as top-$N$ accuracy: a PGPS method is considered correct when the ground truth answer appears within its $N$ candidate answers.
For multiple-choice tasks, the metric depends on the output representation of the method. If the method predicts an option label, evaluation reduces to standard top-$1$ accuracy. If it produces a value, e.g., scalar, a modified version of top-$N$ accuracy is utilized: the $N$ generated values are scanned in order, and the attempt is scored correct once the first value that coincides with any listed option matches the ground truth.

\paragraph{Benchmarks}
Most PGPS benchmarks have been proposed to evaluate model performance on direct-answer and multiple-choice tasks. Some benchmarks exclusively consist of plane geometry problems~\citep{GeoShader, geos, intergps, geoqa, geoqa+, pgps9k, formalgeo7k, TrustGeoGen, geomverse, geosense}, while others include plane geometry problems as part of broader benchmarks designed for general multi-modal reasoning evaluation~\citep{mathvista, mathverse, mmmu, visonlyqa, Math-Vision, dynamath, polymath, vcbench}.

\subsubsection{Reasoning tasks}
\paragraph{Task description}
Some PGPS benchmarks assess methods not only on the correctness of the final answer but also on the soundness of the intermediate reasoning~\citep{unigeo,GPSM4K}. In a widely adopted proving problem setting, a PGPS method must generate a sequence of geometric axioms and theorems that derive the target statement, e.g., two angles are congruent, directly from the given conditions.

\paragraph{Evaluation metrics}
For reasoning-step construction tasks, top-$N$ accuracy is again adopted, granting success when any of the $N$ predicted reasoning steps exactly reproduces the ground-truth steps.

\paragraph{Benchmarks}

UniGeo~\citep{unigeo} is currently the only benchmark designed explicitly to systematically measure reasoning capabilities. 
Recently, approaches leveraging LLMs have emerged to evaluate individual reasoning steps~\citep{mathverse,GPSM4K}.
However, these methods inherently rely on LLMs, posing significant limitations. 
Consequently, proposing diverse and systematic reasoning benchmarks remains an open research challenge.

\section{Overall approach}\label{sec:overall}

\begin{figure*}[t!]
    \centering
    \includegraphics[width=.9\linewidth]{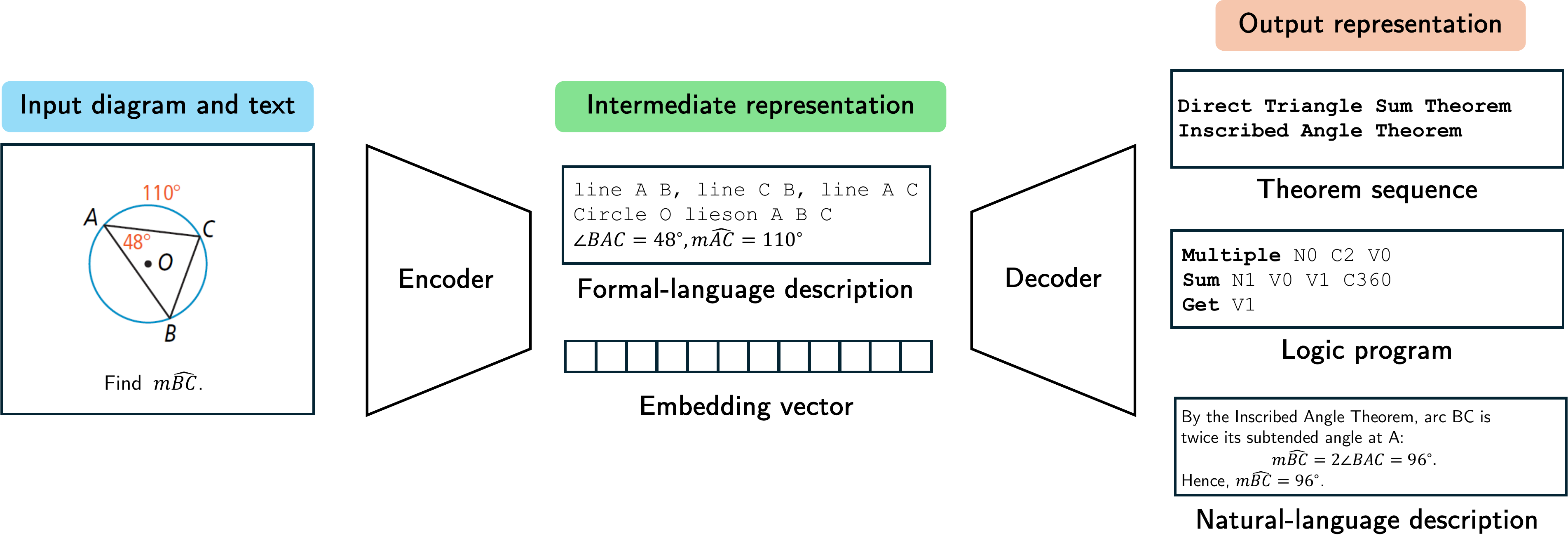}
    \caption{Visualization of the overall structure of PGPS methods. PGPS methods first encode the input diagram and text into an intermediate representation. The encoded representation is then passed to the decoder, which generates the final solution as a theorem sequence, a logic program, or a natural-language description.}
    \label{fig:pgps_architecture}
    \vskip -0.19in
\end{figure*}

PGPS models typically employ an encoder-decoder architecture, as demonstrated in \cref{fig:pgps_architecture}. The \emph{encoder} jointly processes the diagram and textual description to produce an \emph{intermediate representation} that captures essential geometric information of the problem. The \emph{decoder} then utilizes the extracted intermediate representation to generate a solution, presented as either a theorem sequence, a logic program, or a natural-language description. Finally, the answer is obtained by post-processing the generated solution, e.g., by executing the logic program or extracting the final result from the natural-language description.

Before we discuss the detailed approach to constructing the encoder and decoder, we first review the output formats of the encoder and decoder commonly used across different PGPS tasks.

\subsection{Encoder outputs}

The output of an encoder forms an intermediate representation that can be further used as an input to a decoder. We categorize the output format of the encoder into i) formal-language description and ii) embedding vectors.

\paragraph{Formal-language description}

Several studies explicitly extract geometric primitives and relations from given diagram-text pairs, converting them into formal-language descriptions.
A formal-language description consists of an \emph{entity} set and a \emph{predicate} set. 
The entity set contains geometric primitives, e.g., elementary primitives such as points, lines, and circles~\citep{pgdp5k,pgps9k}, or higher-level shapes such as triangles and squares~\citep{geos,geos++,GEOS-OS,intergps}, along with non-geometric tokens such as numbers and variable names. The predicates define the relationships among the entities. For instance, an equality predicate binds two entities $\angle ABC$ and $30^\circ$ to represent the numerical value of the angle, i.e., $\angle ABC = 30^\circ$ or specify geometric relations, such as segments $AB$ and $BC$ being perpendicular, i.e., $AB \perp BC$.

In earlier studies, rule-based approaches~\citep{textparser1,textparser2} and semantic parsers~\citep{semantictextparser} have been proposed to extract formal-language descriptions from textual descriptions without analyzing the diagram~\citep{geos,intergps}. 
Recent works extend these approaches to extract a formal language description from a diagram-text pair.
Consequently, many PGPS studies release datasets consisting of diagrams and formal-language description pairs to train diagram parsers in a supervised way~\citep{geos,geos++,GEOS-OS,pgdp5k,intergps,pgps9k,formalgeo7k}

\paragraph{Embedding vectors}

Certain PGPS encoders represent inputs as embedding vectors, typically utilizing one of three strategies: i) embedding diagrams and textual descriptions separately and subsequently merging them~\citep{geoqa,geoqa+,unigeo,sca-gps,unimath,FLCL-GPS}, ii) embedding diagrams exclusively and then combining them with raw textual inputs~\citep{geox,GeoDANO,math-llava,mavis,G-LLaVA,DFE-GPS,Chimera,Geo-LLaVA}, or iii) jointly processing diagrams and texts through a unified encoder~\citep{pgps9k,LANS}. 
Although these embeddings are generally less interpretable compared to formal-language descriptions, they enable end-to-end training with the decoder.

\subsection{Decoder outputs}

Given the output of the encoder, the decoder generates the solution from which the final answer can be derived. We classify decoder output formats into three types: i) theorem sequences, ii) logic programs, and iii) natural-language descriptions.

\paragraph{A sequence of theorems}
Many PGPS works represent the output of a PGPS problem as a sequence of theorem applications.
This approach naturally aligns with a reasoning process, in which theorems are iteratively applied to given entities and predicates to logically derive new geometric facts, including the target predicate specified as a goal~\citep{alphageometry}. Specifically, given geometric entities and predicates extracted from the original description, theorems from a predefined library can be applied to the entities and predicates to derive additional predicates not explicitly stated in the original problem. Recent PGPS datasets provide annotated triples consisting of the formal-language description, the target predicate, and a corresponding reference theorem sequence~\citep{geos++,GEOS-OS,intergps,formalgeo7k}.

\paragraph{A logic program}

A logic program is commonly adopted as an output representation for PGPS. 
Specifically, inspired by the observation that the reasoning process in PGPS typically involves applying a series of operations to numerical constants and variables provided in the problem~\citep{geoqa,mathqa,PoT}, a logic program is defined as a sequence of triples, each consisting of an operation and its operands, such as numerical values and variable names. 
The operations in these programs fall into two main categories: i) arithmetic functions, ranging from basic operations like addition and multiplication to geometry-specific computations such as the Pythagorean operation~\citep{geoqa,geoqa+,unigeo}, and ii) equality assertions that establish identity between two expressions~\citep{pgps9k}. 
Several PGPS datasets provide paired examples, each consisting of a diagram-text problem and its corresponding logic program~\citep{geoqa, geoqa+, unigeo, pgps9k}.

\paragraph{A natural-language description}

Recent PGPS methods generate solutions and answers in natural language without relying on a specific template. The inherent flexibility of natural language allows these models to easily provide outputs for a wide range of tasks, e.g., geometric diagram captioning, without being limited to fixed problem-solving formats.
To train such methods, various types of PGPS datasets have been proposed. For tasks which focus on problem solving, the output, given a diagram and text, can either be the answer expressed in natural language~\citep{math-llava} or a reasoning path in the form of a chain-of-thought~\citep{CoT} to infer the answer~\citep{mavis,G-LLaVA}. In addition to problem solving, datasets have also been proposed for tasks such as geometric diagram captioning~\citep{mavis,G-LLaVA,GeoDANO,geox} and question answering~\citep{G-LLaVA}.

\begin{figure*}[t!]
    \centering
    \includegraphics[width=.9\linewidth]{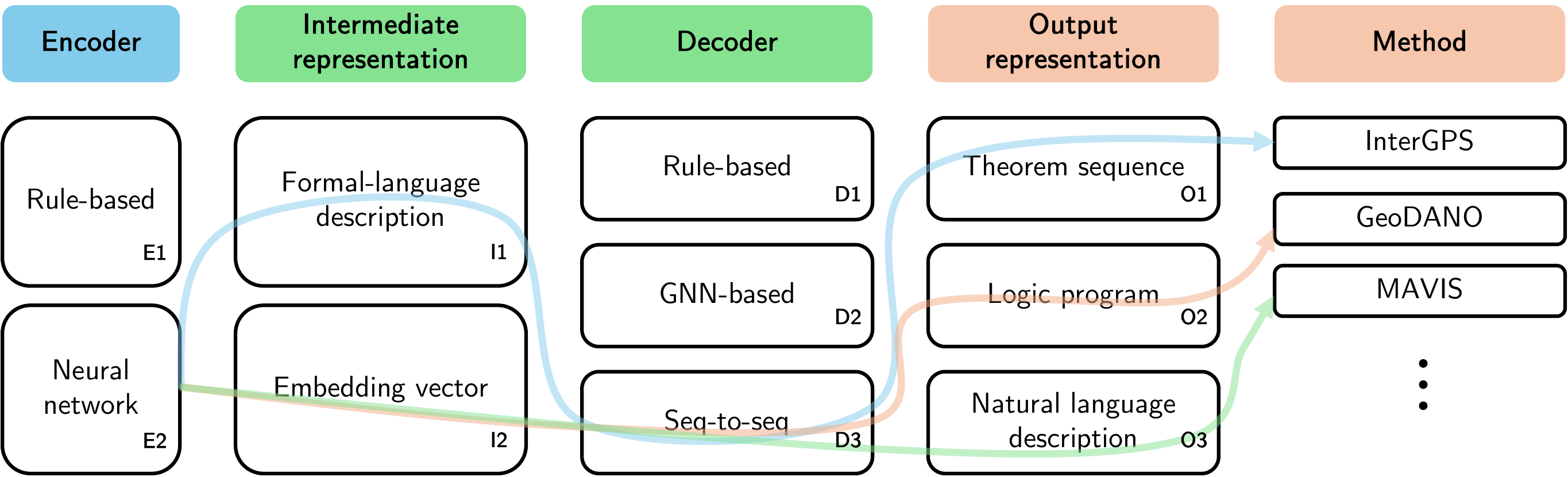}
    \caption{Overview of the PGPS pipeline. PGPS methods can be categorized based on the combination of the encoder, intermediate representation, decoder, and output representation. For example, the InterGPS can be represented as a combination of E2, I1, D3, and O1.
    We summarize PGPS methods as a combination of these components in \cref{tab:pgps_methods}. }
    \label{fig:pgps_methods}
\end{figure*}

\subsection{Encoder-decoder with desired outputs}
Once the intermediate representations and output representations are determined based on target problems or tasks, one can choose an appropriate encoder and decoder that can produce the desired outputs. \cref{fig:pgps_methods} summarizes possible combinations of encoder-decoder architectures along with the desired outputs. A combination of encoder, intermediate representation, decoder, and output representation can lead to a specific architecture for PGPS. In the following two sections, we review the possible choices of encoder and decoder structures.

\section{Encoders}\label{sec:encoder}

The encoder extracts the relevant components from the given diagram and text that are necessary for PGPS.
We review the encoders in the following aspects: i) rule-based and ii) neural network-based.

\subsection{Rule-based encoders}

Early PGPS methods relied on classical computer vision and text parsing techniques to independently extract geometric primitives and relations from diagrams and text, merging them into formal-language descriptions. 
Most studies~\citep{geos,geos++,GEOS-OS,GeoShader,S2} employed rule-based diagram parsers, notably HoughGeo~\citep{HoughGeo} or G-Aligner~\citep{G-Aligner}, which preprocess diagrams to detect geometric primitives using classical computer vision techniques, e.g., Gaussian blur and Hough transforms, and then match detected primitives to literal sets using either handcrafted rules or optimization. 
For textual extraction, many approaches~\citep{E-GPS,Pi-GPS,geodrl,FGeo-HyperGNet,GCN-GPS,FGeo-DRL} adopted the InterGPS~\citep{intergps} parser, a rule-based method utilizing regular expressions, which is reliable and effective even with limited data. 

\subsection{Neural network encoders}
We review the neural network-based encoders based on the desired output format.

\subsubsection{Formal-language description generation}

Recent PGPS approaches adopt neural encoders to generate formal-language descriptions from diverse diagrams and texts, typically training separate encoders for each modality. 
Neural diagram encoders commonly operate in two stages: primitive detection using object detectors such as RetinaNet~\citep{retinanet, intergps} and feature pyramid networks~\citep{fpn,pgdp5k}, followed by relation inference modeled either as a constrained optimization problem~\citep{intergps} or as a graph-learning task leveraging graph neural networks (GNNs)~\citep{pgdp5k}. 
For text encoding, subsequent PGPS studies~\citep{geos++,GEOS-OS} commonly employ logistic regression models, as originally introduced by GEOS~\citep{geos}, to extract primitives and relations from problem statements.




\subsubsection{Embedding vector generation}

To enable end-to-end learning, recent PGPS methods employ neural encoders that map both the diagram and text into a unified embedding space, providing a joint vector representation for PGPS. Here, we review the neural encoders based on their training strategy.

\paragraph{Learning from scratch}
 
Early PGPS works train joint diagram-text encoders and decoders end-to-end from scratch on target PGPS datasets.
Diagram embeddings commonly utilize convolutional neural networks (CNNs), including vanilla CNN~\citep{pgps9k}, ResNet~\citep{resnet, geoqa,geoqa+}, DenseNet~\citep{densenet, GCN-GPS}, and VQ-VAE encoders~\citep{vq-vae,unimath}, as well as Vision Transformers (ViT)~\citep{vit,sca-gps}. 
Text embeddings are typically produced by sequential models like LSTMs~\citep{lstm,geoqa, geoqa+} or Transformer-based encoders~\citep{transformer}, such as vanilla Transformer~\citep{pgps9k, sca-gps, LANS} and RoBERTa~\citep{roberta,geoqa+}. 
Diagram and text embeddings are fused via co-attention networks~\citep{co-attention,geoqa,sca-gps}, bi-directional GRUs~\citep{gru,pgps9k,LANS}, or Transformers~\citep{unigeo}.

Besides direct optimization on PGPS tasks, joint encoders frequently employ auxiliary objectives for improved performance. 
Many approaches incorporate self-supervised tasks, including jigsaw-location prediction~\citep{geoqa,geoqa+, GCN-GPS}, masked-token prediction in text~\citep{bert,unigeo, pgps9k, LANS} or diagrams~\citep{mae,sca-gps}, text-conditioned diagram-symbol classification~\citep{sca-gps}, and VQ-VAE objective~\citep{unimath}. 
Other studies leverage explicit labels, training encoders for geometry-element or knowledge-point classification~\citep{geoqa,geoqa+}, or contrastive learning between diagram patches and textual tokens~\citep{LANS}.



\paragraph{Pre-trained encoders}

To leverage pretrained knowledge and enhance training efficiency, many recent PGPS methods employ neural encoders inspired by the LLaVA architecture~\citep{llava}, which integrates a pretrained vision encoder to encode diagrams.
Specifically, diagrams are first transformed into visual embeddings using a pretrained vision encoder, followed by a lightweight adapter consisting of a multi-layer perceptron. 
During training, only the adapter parameters are updated, keeping the vision encoder frozen to preserve general visual knowledge and reduce training cost. 
While OpenCLIP~\citep{clip} is the most commonly used backbone~\citep{math-llava,G-LLaVA,Geo-LLaVA}, other general-purpose models such as SigLIP~\citep{siglip,DFE-GPS} and InternViT~\citep{internvl,Chimera}, as well as the math-specific Math-CLIP encoder~\citep{mavis,Chimera}, have also been employed.


\paragraph{Fine-tuned encoders}

Most pretrained vision encoders perform poorly when applied to geometric diagrams~\citep{mavis,geox,GeoDANO}. To address this limitation, PGPS methods employing the LLaVA-style architecture typically fine-tune the vision encoders before integrating them into downstream pipelines. Two main fine-tuning strategies are common: i) self-supervised methods such as masked auto-encoding~\citep{mae,geox}, and ii) weakly supervised methods such as CLIP~\citep{mavis,GeoDANO}, direct preference optimization~\citep{dpo,CogAlign}, or grounding tasks~\citep{glip,sve-math}, which leverage synthetic geometric diagrams and labels pairs.
Nevertheless, since synthetic diagrams do not fully capture the characteristics of real-world diagrams, GeoDANO~\citep{GeoDANO} further employs few-shot domain adaptation under the same CLIP training objective to minimize the residual domain gap.

\section{Decoders}\label{sec:decoder}

Based on the representations produced by the encoder, the decoder generates the solution to the problem.
We survey the PGPS decoders using the following dimensions: i) input representation and ii) architectural design.

\subsection{Formal-language description decoder}

We first introduce the architectures of the decoders that receive a formal language description as input.

\paragraph{Rule-based axiomatic decoders}

Several methods that operate on formal-language descriptions determine the required theorem sequence with a rule-based decoder. GEOS++~\citep{geos++} employs an exhaustive brute-force search to locate a sequence of theorems whose application yields the target predicate. GeoShader~\citep{GeoShader} specifies a deterministic set of composition rules that directly selects the relevant theorems without search. GEOS-OS~\citep{GEOS-OS} trains a log-linear model to assign probabilities to candidate theorems and then performs beam search, returning the highest-scoring theorem sequence.

\paragraph{GNN-based decoders}

A formal-language description, composed of geometric primitives and their relations, naturally corresponds to a graph structure. 
Exploiting this, several PGPS decoders first encode the formal description as a graph or hypergraph and then generate theorem-application sequences from the resulting graph representation. 
Such encodings typically follow one of three schemes: i) primitives as nodes and predicates as edges~\citep{geodrl}, ii) primitives and predicates both as nodes connected via edges~\citep{GCN-GPS}, or iii) predicates as hypernodes and theorems as directed hyperedges forming a hypertree~\citep{FGeo-HyperGNet}. 
These encoded structures are subsequently fed into graph-to-sequence decoders, such as Graphormer~\citep{FGeo-HyperGNet}, graph Transformer~\citep{geodrl}, or graph convolutional network~\citep{gcn} followed by LSTM~\citep{GCN-GPS}, to produce the target theorem sequence.



\paragraph{Sequence-to-sequence decoders}

Some approaches treat formal-language descriptions as a flat token sequence and pass it directly to a sequence-to-sequence (seq-to-seq) model to generate the corresponding theorem sequence. Transformers are predominantly employed for these tasks by encoding the formal description directly~\citep{intergps,E-GPS,FGeo-DRL}. 
A few studies instead utilize off-the-shelf LLMs, e.g., o3-mini~\citep{o3-mini}, without additional training~\citep{Pi-GPS}.

\subsection{Seq-to-seq embedding decoders}

Several PGPS studies feed either a joint diagram–text embedding or a concatenation of diagram embedding and raw text into a sequence-to-sequence decoder. 
Early work primarily employs RNN-based decoders such as LSTMs or GRUs~\citep{geoqa,geoqa+,pgps9k,LANS,sca-gps,FLCL-GPS}, while later studies commonly adopt encoder–decoder Transformers such as T5~\citep{t5, unimath,unigeo}. 
The recent proliferation of LLMs has motivated a shift toward fine-tuning encoder-only Transformers, such as LLaMA~\citep{llama,GeoDANO,G-LLaVA,Geo-LLaVA} and Vicuna~\citep{vicuna,math-llava}, specifically adapted for PGPS tasks. 
\begin{table*}[ht]
\centering
\resizebox{\linewidth}{!}{
\begin{tabular}{p{0.15\textwidth} p{0.26\textwidth} p{0.26\textwidth} p{0.26\textwidth}}
\toprule
 & \textbf{Example 1} & \textbf{Example 2} & \textbf{Example 3}\\
\midrule
\textbf{Diagram} &
\begin{minipage}[t]{\linewidth}\vspace{0pt}\centering
  \includegraphics[width=\linewidth]{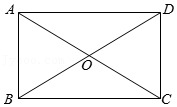}
\end{minipage} &
\begin{minipage}[t]{.7\linewidth}\vspace{0pt}\centering
  \includegraphics[width=\linewidth]{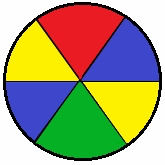}
\end{minipage} &
\begin{minipage}[t]{.7\linewidth}\vspace{0pt}\centering
  \includegraphics[width=\linewidth]{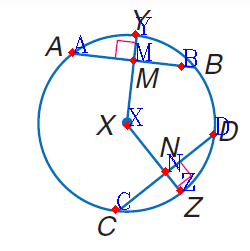}
\end{minipage} \\
\midrule
\textbf{Question} &
\small In the given figure, let's denote the area of triangle AOB as variable x. Find the area of rectangle ABCD in terms of x. \textbf{Choices: A: 8 B: 10 C: 12 D: 16} &
\small Based on the image, what is the measure of the interior angle at \textbf{vertex A}? Choices: A. 90 degrees B. More than 90 degrees C. Less than 90 degrees D. Cannot be determined &
\small Does the diagram include any line segments that are not perpendicular to each other? \\
\midrule
\textbf{Solution} &
\small To determine the area of rectangle ABCD, we can use the fact that triangle AOB is half the area of the rectangle. \textbf{Therefore, the area of rectangle ABCD is 2 times the area of triangle AOB, which is 2x.} Hence, the answer is \textbf{option B. Answer:D} &
\small Use the properties of the geometric shapes and theorems related to angles to deduce the measure of \textbf{the interior angle at vertex A} based on the given image and information. So the \textbf{answer is B} &
\small Yes, in the diagram, \textbf{line segment YM is not perpendicular to line segment MA}. \\
\bottomrule
\end{tabular}
}
\caption{Examples of hallucinations in the natural-language description datasets annotated with L(V)LM. We visualize the examples from the PGPS datasets, e.g., G-LLaVA and MAVIS, which contain hallucinations in the question or response due to the L(V)LM annotation. We highlight the hallucinations with bold characters.}
\label{tab:dataset_hallucinations}
\end{table*}

\section{Challenges and future directions}\label{sec:challenges}

We examine the remaining challenges in PGPS and propose potential directions for future work.

\subsection{Hallucination in diagram perception}\label{sec:challenge_hallucination}

PGPS methods initially extract geometric primitives and relations from diagrams and text, making accurate perception crucial before reasoning. However, studies indicate that PGPS methods frequently misperceive these primitives and relations, especially when generating natural-language descriptions~\citep{CogAlign,mathverse} as depicted in \cref{fig:perception_error}. For example, \cref{tab:perception_error} reveals that GPT-4.1~\citep{gpt-4.1} fails to capture a fundamental geometric relation among the points and lines and produces hallucinations. 
These hallucinations not only degrade PGPS performance but also diminish dataset quality. Computer vision studies report similar hallucination issues in datasets produced by large VLMs~\citep{hallucination1,hallucination2,hallucination3,hallucination4}, further evidenced in PGPS datasets as shown in \cref{tab:dataset_hallucinations}. Consequently, models trained on hallucinated data suffer measurable performance declines~\citep{hallucination1,hal_effect1,hal_effect2,hal_effect3}.

Visual prompting techniques, such as augmenting diagrams with bounding boxes, markers, or segmentation masks, have emerged as promising solutions for mitigating hallucinations~\citep{visualprompting1,visualprompting2,visualprompting3}. These methods are especially beneficial for PGPS tasks, as they dynamically highlight relevant primitives and relations during reasoning and facilitate the critical step of drawing auxiliary lines. Augmenting diagrams at test time~\citep{s1} by applying segmentation masks~\citep{sam2} or adding auxiliary constructions aligned with the current reasoning step~\citep{diagrammatic,visualsketchpad} offers a practical approach to enhance multi-modal reasoning performance in PGPS.

\subsection{Evaluation challenges in benchmarks}\label{sec:challenge_comprehensive}

Comprehensive PGPS benchmarks should evaluate perception across diverse, realistic diagrams, ensuring that visual processing is essential for solving each problem. However, as shown in \cref{tab:comprehensive_benchmark}, existing benchmarks do not satisfy these criteria simultaneously. Synthetic diagrams, while scalable, often fail to represent the complexity of real-world scenarios~\citep{synthetic1,synthetic2,synthetic3}, lacking elements such as parallel markers or placeholder objects, as illustrated in \cref{fig:synthetic_diagrams}. Conversely, manually collected benchmarks better reflect real-world complexity but frequently reuse diagrams from popular PGPS datasets, introducing data leakage and compromising domain generalization evaluations~\citep{leakage1,leakage2,leakage3}.

Even manually curated benchmarks without common PGPS dataset reuse often neglect crucial diagram–text dependencies discussed in \cref{sec:diagram_dependency}. MathVerse addresses these dependencies explicitly and avoids synthetic diagrams, but still suffers from data leakage, limiting its capability to assess genuine multi-modal reasoning. To overcome these issues, future research should develop synthetic diagram generators that closely replicate real-world complexity or create new datasets that strictly require visual reasoning while rigorously preventing data leakage.

\section{Conclusion}

In this paper, we examine the tasks, benchmarks, and methods used in existing PGPS research. 
We summarize the main PGPS approaches as an encoder-decoder architecture, along with the intermediate and output representations utilized across different methods
Through the analysis, we outline future research directions addressing current challenges, particularly regarding diagram perception and benchmark comprehensiveness.
\clearpage
\section*{Limitations}

In this paper, we primarily survey studies related to PGPS. While our work offers a comprehensive review of the existing PGPS literature, it is limited to two-dimensional geometry. Consequently, we do not address research involving three-dimensional geometry, such as projective and solid geometry, which requires understanding spatial relationships in three-dimensional space.

\bibliography{ref}

\clearpage
\appendix

\renewcommand{\thefigure}{A\arabic{figure}}
\setcounter{figure}{0}
\renewcommand{\thetable}{A\arabic{table}}
\setcounter{table}{0}

\section{Additional axis on benchmark dataset}\label{sec:misc}

\subsection{Reasoning complexity}

We discuss the mathematical concepts and difficulty levels encountered in plane geometry problems used by existing benchmarks and datasets. 
Typical plane geometry problems involve calculating specific angle measures, arc measures, segment or arc lengths, and areas of designated regions. 
Computing these numerical values generally requires basic arithmetic and root operations, but may also involve trigonometric functions, such as sine and cosine. 
Although no standardized quantitative method currently exists to measure problem difficulty, problems can be qualitatively categorized according to their original sources, such as SAT exams~\citep{geos,geos++,GEOS-OS}, plane geometry curricula from grades 6–12 American~\citep{intergps,pgps9k,mm-math} or Chinese school~\citep{geoqa,geoqa+,geosense}, college-level mathematics~\citep{mmmu}, or mathematics competitions, e.g., AMC 8, 10, and 12~\citep{Math-Vision}.

\subsection{Diagram-text redundancy}\label{sec:diagram_dependency}

To serve as rigorous benchmarks and datasets for multi-modal reasoning, the collected problems must require simultaneous interpretation of both diagrams and accompanying textual descriptions. By contrast, PGPS problems that can be solved using the text alone cannot effectively evaluate the diagram-text integration capability of PGPS methods. Nevertheless, many existing benchmarks and datasets still contain such problems, thereby inadequately assessing the perception abilities of PGPS methods~\citep{mathverse}.

Recent PGPS benchmarks have addressed this limitation by explicitly annotating problems with modality-specific information and subsequently removing redundant textual cues~\citep{intergps, pgps9k, mathverse}. Several benchmarks provide multiple variants of each problem for more fine-grained analysis of diagram-text dependency. For instance, MathVerse~\citep{mathverse} relocates selected information from the text into the diagram, while DynaMath~\citep{dynamath} generates alternative diagrams and corresponding answers based on a single textual description. Thus, failure to solve certain variants of the same problem indicates that the model is not genuinely utilizing the diagram.

\subsection{Data collection methods}
We summarize three data collection methods mainly used to construct PGPS datasets.

\paragraph{Human annotation}

In most cases, datasets are constructed through human annotation based on problems sourced from textbooks, internet sites, or similar resources~\citep{geos,geoqa,intergps,mathvista,mm-math,mmmu}. This involves manually collecting problems and having human annotators provide the corresponding outputs. Additionally, some studies apply text augmentation techniques, such as back-translation, to diversify the text style and enrich the dataset~\citep{geoqa+}.

\paragraph{Synthetic annotation}

Several PGPS studies create synthetic benchmarks and datasets instead of collecting problems from textbooks or the internet. 
These studies typically implement synthetic engines to generate diagrams and corresponding structured information. 
For example, synthetic engines can generate captions containing the geometric information explicitly present in diagrams~\citep{mavis}, or use symbolic reasoning engines to produce reasoning steps that derive the stated goals from diagram-text pairs~\citep{mavis,geomverse,TrustGeoGen}. 
Such synthetic approaches offer clear advantages, including easy scalability and guaranteed completeness of annotations. 
However, they often struggle to produce sufficiently diverse diagrams that accurately reflect the real-world problems. This limitation is further discussed in \cref{sec:challenge_comprehensive}.


\paragraph{L(V)LM-assisted annotation}

For certain datasets, particularly those with natural-language description as the output representation, LLMs and VLMs such as GPT~\citep{gpt} or GPT-4V~\citep{gpt-4v} are employed for dataset construction. Specifically, problems and solutions are sourced from datasets like GeoQA+, UniGeo, or PGPS9K, and GPT or GPT-4V are used to augment these by generating multiple problem-solution pairs for a given problem scenario~\citep{G-LLaVA,math-llava,mavis}. Alternatively, some studies apply the same process to synthetic data, such as diagram-caption pairs generated by a synthetic data engine~\citep{mavis,geomverse}.
However, due to the poor perception ability of GPT-4V, several hallucinations occur in the augmented datasets. We discuss more details about the challenge in \cref{sec:challenge_hallucination}.

\section{PGPS Methods}

\subsection{Summary of PGPS methods}

We summarize the PGPS methods in terms of the encoder, intermediate representation, decoder, and the output format at \cref{tab:pgps_methods}.

\begin{table*}[t!]
\centering
\begin{tabular}{ccccp{8cm}}
\toprule
Encoder & Intermediate & Decoder & Output & \multicolumn{1}{c}{Methods} \\ \midrule
E1 & I1 & -- & -- & HoughGeo~\citep{HoughGeo}, \; G-Aligner~\citep{G-Aligner}, \; GEOS~\citep{geos} \\ \addlinespace[2pt]
E2 & I1 & -- & -- & PGDPNet~\citep{pgdp5k}, \;FGeo\textendash Parser~\citep{FGeo-Parser} \\ \addlinespace[2pt]
E1 & I1 & D1 & O1 & GEOS++~\citep{geos++}, \;GEOS\textendash OS~\citep{GEOS-OS}, \;GeoShader~\citep{GeoShader}, \;S2~\citep{S2} \\ \addlinespace[2pt]
E2 & I1 & D2 & O1 & FGeo\textendash HyperGNet~\citep{FGeo-HyperGNet}, \;GCN\textendash GPS~\citep{GCN-GPS}, \; GeoDRL~\citep{geodrl} \\ \addlinespace[2pt]
E2 & I1 & D3 & O1 & InterGPS~\citep{intergps}, \;E\textendash GPS~\citep{E-GPS}, \;Pi\textendash GPS~\citep{Pi-GPS}, \;FGeo\textendash DRL~\citep{FGeo-DRL} \\ \addlinespace[2pt]
E2 & I1 & D1 & O1 & FGeo\textendash SSS~\citep{fgeo-sss} \\ \addlinespace[2pt]
E2 & I1 & D2 & O3 & GOLD~\citep{GOLD} \\ \addlinespace[2pt]
E2 & I2 & D3 & O2 & NGS~\citep{geoqa}, \;DPE\textendash NGS~\citep{geoqa+}, \;Geoformer~\citep{unigeo}, \;PGPSNet~\citep{pgps9k}, \;SCA\textendash GPS~\citep{sca-gps}, \;UniMath~\citep{unimath}, \;FLCL\textendash GPS~\citep{FLCL-GPS}, \;LANS~\citep{LANS}, \;GeoX~\citep{geox}, \;GeoDANO~\citep{GeoDANO} \\ \addlinespace[2pt]
E2 & I2 & D3 & O3 & Math\textendash LLaVA~\citep{math-llava}, \;Visual~SKETCHPAD~\citep{visualsketchpad}, \;MAVIS~\citep{mavis}, \;G\textendash LLaVA~\citep{G-LLaVA}, \;DFE\textendash GPS~\citep{DFE-GPS}, \;Chimera~\citep{Chimera}, \;Geo\textendash LLaVA~\citep{Geo-LLaVA}, SVE-Math~\citep{sve-math} \\ \bottomrule
\end{tabular}
\caption{Categorization of existing PGPS methods. We categorize the PGPS methods based on their encoder, intermediate representation, decoder, and output format. The symbols come from \cref{fig:pgps_methods}.}
\label{tab:pgps_methods}
\end{table*}

\section{Challenges and Future Directions}

\subsection{Error analysis on wrong responses}

\begin{figure}[h!]
    \centering
    \includegraphics[width=.9\linewidth]{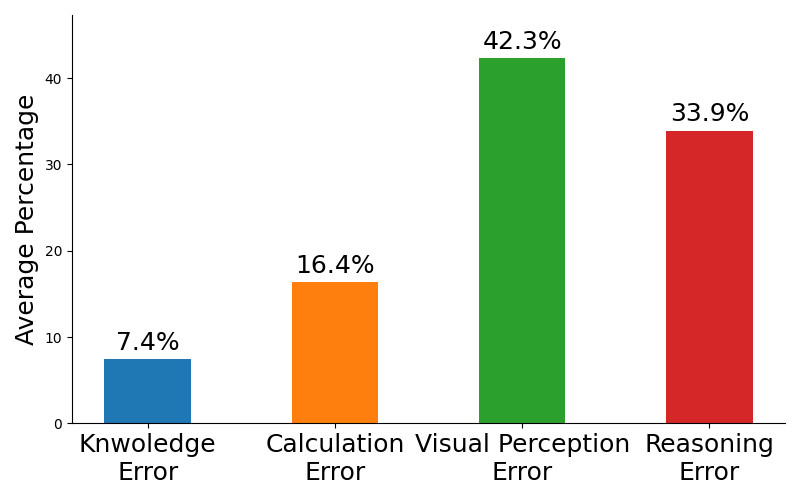}
    \caption{Error analysis on the response of GPT-4V on MathVerse. We analyze the responses of GPT-4V on MathVerse, reporting the average percentage for each type of error across five MathVerse variants, Text Dominant, Text Lite, Vision Intensive, Vision Dominant, and Vision Only, which are reported in MathVerse. Our analysis indicates that incorrect answers predominantly result from visual perception and reasoning errors.}
    \label{fig:perception_error}
\end{figure}

\subsection{Examples of perception hallucinations}

We provide examples of hallucinated responses by GPT-4.1 in \cref{tab:perception_error}.

\begin{table*}[t!]
\centering
\resizebox{\linewidth}{!}{
\begin{tabular}{p{0.20\textwidth} p{0.38\textwidth} p{0.38\textwidth}}
\toprule
 & \textbf{Example 1} & \textbf{Example 2}\\
\midrule
\textbf{Input prompt} &
\multicolumn{2}{p{\dimexpr0.76\textwidth}}{In the given diagram, extend the segment SU to form an infinite straight line. This line divides the two-dimensional plane into two half-planes. The points Q and A are also shown in the diagram. Are the points Q and A located on the same side of the line SU?} \\
\midrule
\textbf{Input diagram} &
\begin{minipage}[t]{\linewidth}\vspace{0pt}\centering
  \includegraphics[width=\linewidth]{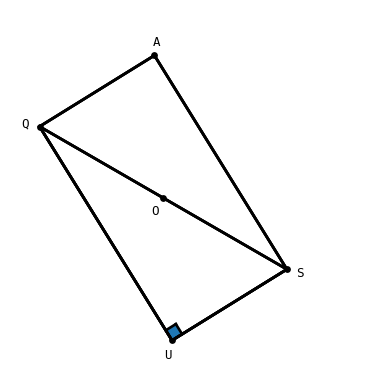}
\end{minipage} &
\begin{minipage}[t]{\linewidth}\vspace{0pt}\centering
  \includegraphics[width=\linewidth]{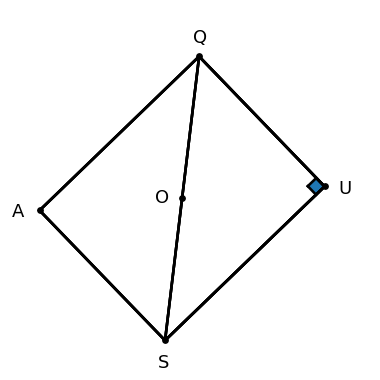}
\end{minipage} \\
\midrule
\textbf{Response} &
To determine if points Q and A are on the same side of the line SU, observe the diagram: extending SU clearly shows that point Q is on one side of the line, \textbf{while point A is on the opposite side}. The quadrilateral QASU suggests \textbf{A and Q are separated by the segment SU itself.} & 
To determine whether points Q and A are on the same side of the line SU, we observe the diagram. The segment SU is extended to form a straight line. Point Q is above SU, \textbf{while point A is on the opposite side}, to the left of SU. Therefore, \textbf{the points Q and A are on opposite sides of line SU.}
\\
\bottomrule
\end{tabular}
}
\caption{Examples of hallucinations occured in GPT-4.1 responses. We analyzed the diagram perception performance of GPT-4.1~\citep{gpt-4.1}, specifically determining whether two points are on the same side of a given line. We generated 100 problems using the synthetic data engine from GeoDANO~\citep{GeoDANO} and tested them with GPT-4.1, observing a low accuracy of 59\%. The examples above illustrate cases where GPT-4.1 hallucinated and provided incorrect answers. Hallucinated parts are highlighted in bold. }
\label{tab:perception_error}
\end{table*}

\newpage
\subsection{Comprehensivity of current PGPS benchmarks}

\begin{table}[H]
\centering
\resizebox{\linewidth}{!}{
\begin{tabular}{lccc}
\toprule
\textbf{Methods}&
\textbf{\makecell{Realistic styles\\of diagrams}} &
\textbf{\makecell{No data\\leakage}} &
\textbf{\makecell{Diagram--text\\interdependence}}\\
\midrule
MMMU          & $\bigcirc$ & $\bigcirc$ & $\times$ \\
Math\textendash V   & $\bigcirc$ & $\bigcirc$ & $\times$ \\
MathVista     & $\bigcirc$ & $\times$  & $\times$ \\
MathVerse     & $\bigcirc$ & $\times$  & $\bigcirc$ \\
GeomVerse     & $\times$   & $\bigcirc$ & $\times$ \\
VisOnlyQA     & $\times$   & $\bigcirc$ & $\bigcirc$ \\
MM\textendash Math   & $\bigcirc$ & $\bigcirc$ & $\times$ \\
GeoEval       & $\times$   & $\times$  & $\times$ \\
DynaMath      & $\bigcirc$ & $\times$  & $\bigcirc$ \\
\bottomrule
\end{tabular}

}
\caption{Comprehensivity across existing PGPS benchmarks. The table summarizes benchmark features in terms of realistic diagram styles, absence of data leakage, and consideration of diagram-text interdependence.}
\label{tab:comprehensive_benchmark}
\end{table}

\subsection{Synthetic and real-world geometric diagrams}

\begin{figure}[h!]
    \centering
    \begin{subfigure}[t]{0.45\linewidth}
        \centering
        \includegraphics[width=\linewidth]{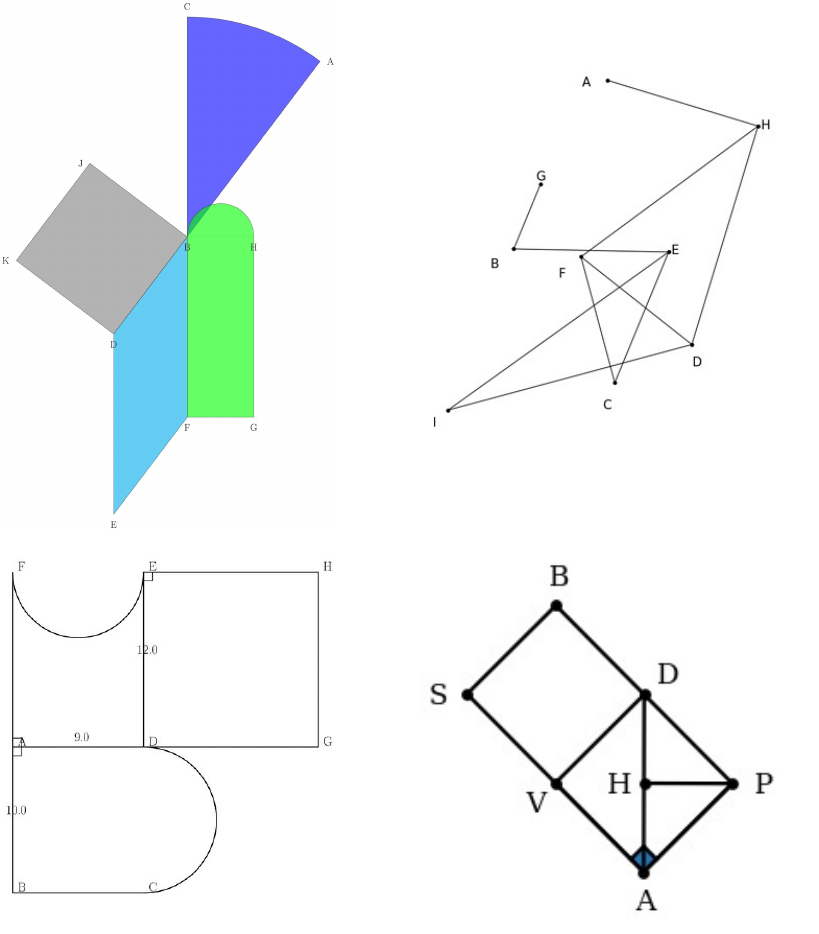}
        \caption{Synthetic diagrams}
    \end{subfigure}
    \begin{subfigure}[t]{0.45\linewidth}
        \centering
        \includegraphics[width=\linewidth]{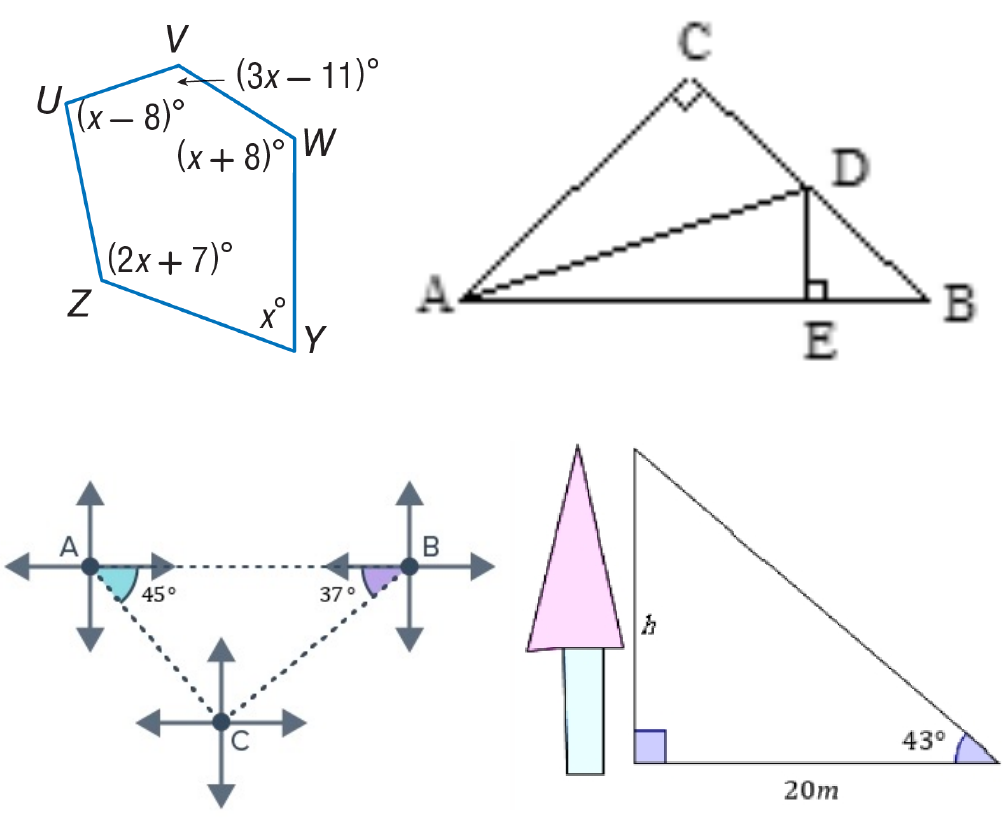}
        \caption{Real-world diagrams}
    \end{subfigure}
    \caption{Visualization of the synthetic and real-world geometric diagrams. We compare the geometric diagrams, which are synthetically generated or manually collected from existing sources. The synthetic diagrams are from GeomVerse, VisOnlyQA, MAVIS, and GeoDANO. The real-world diagrams are from MathVerse.}
    \label{fig:synthetic_diagrams}
\end{figure}

\end{document}